%% file: main.tex
\documentclass[conference]{IEEEtran}
\IEEEoverridecommandlockouts

\usepackage{cite}
\usepackage{amsmath,amssymb,amsfonts}
\usepackage{algorithmic}
\usepackage{graphicx}
\usepackage{textcomp}
\usepackage{xcolor}
\usepackage{booktabs}
\usepackage{multirow}
\usepackage{marvosym}
\usepackage{float}
\DeclareMathOperator*{\argmin}{arg\,min}
\def\BibTeX{{\rm B\kern-.05em{\sc i\kern-.025em b}\kern-.08em
    T\kern-.1667em\lower.7ex\hbox{E}\kern-.125emX}}
\begin{document}

\title{Learn from Balance: Rectifying Knowledge Transfer for Long-Tailed Scenarios}

\author{
Xinlei Huang$^{1,2}$, 
  Jialiang Tang$^{1}$, 
  Xubin Zheng$^{2}$, 
  Jinjia Zhou$^{3}$, 
  Wenxin Yu$^{1}$, 
  Ning Jiang$^{1}$\thanks{Corresponding author: jiangning@swust.edu.cn} \\
  $^{1}$School of Computer Science and Technology, Southwest University of Science and Technology \\
  $^{2}$School of Information Science and Technology, Great Bay University \\
  $^{3}$Graduate School of Science and Engineering, Hosei University
}


\maketitle
\begin{abstract}
\input{_Abstract}
\end{abstract}


\section{Introduction}
\input{_Intro}


\section{Method}

\input{_Method}

\section{Experiments}
\input{_Experiments}

\section{Conclusion}
\input{_Conclusion}




\end{document}


\title{Supplementary Materials for KRDistill}

\author{
Xinlei Huang$^{1,2}$, 
  Jialiang Tang$^{1}$, 
  Xubin Zheng$^{2}$, 
  Jinjia Zhou$^{3}$, 
  Wenxin Yu$^{1}$, 
  Ning Jiang$^{1}$\thanks{Corresponding author: jiangning@swust.edu.cn} \\
  $^{1}$School of Computer Science and Technology, Southwest University of Science and Technology \\
  $^{2}$School of Information Science and Technology, Great Bay University \\
  $^{3}$Graduate School of Science and Engineering, Hosei University
}
\maketitle

\section{Related work}
\subsection{Knowledge Distillation}
Knowledge distillation aims to train a lightweight and accurate student network by mimicking the informative knowledge of a powerful yet cumbersome teacher network.
Existing knowledge distillation methods can be categorized into three groups based on the type of knowledge transferred by the teacher network, including logit-based, feature-based, and relation-based.
Among them, the logit-based methods~\cite{hinton2015,Zhao2022} propose to soft or decouple the predictions of the teacher network to provide expressive supervision signals to the student network. 
On the other hand, feature-based methods~\cite{Romero2014,Zagoruyko2016,Chen2021} find that high-dimensional features of the teacher network contain more information than low-dimensional logits. 
Therefore, they transfer the meaningful middle-layer attention features or representations from the teacher network to improve the performance of the student network. 
Instead of directly transmitting the logits or features output by the teacher network, relation-based distillation methods~\cite{Tung2019,Park2019} explore instance-level or category-level relations as a form of knowledge. 
As a result, the student network that effectively mimics these relations of the teacher network can produce representations similar to those of the teacher network.

While the aforementioned methods excel in training reliable student networks on balanced standard datasets, they face challenges when applied to real-world imbalanced data. 
This imbalance biases the teacher network towards head categories, leading to suboptimal student network performance.
To mitigate the impact of imbalanced data on the distillation effect, we modify the imbalanced feature representations and logits of the teacher network to enhance the performance of the student network. 

\subsection{Long-Tailed Learning}

Long-tailed learning methods aim to alleviate the issue encountered in data imbalance scenarios, where the model tends to overly focus on the head classes, resulting in poor performance on the tail classes. 
Existing long-tailed learning methods mainly leverage re-sampling, re-weighting, and multi-expert methods to mitigate data imbalances and ensure reliable model performance.
Re-sampling methods provide relatively balanced data to the model by oversampling the tail classes~\cite{Pouyanfar2018} or undersampling the head classes~\cite{Buda2018}.
The re-weighting~\cite{Cui2019,Cao2019} methods enhance the influence of tail class examples on model gradient updates by increasing the weight of tail class examples.
These methods effectively solve the interference of imbalanced data in the model optimization process.

In this paper, we consider the imbalanced knowledge from the teacher network as a crucial supervisory signal for optimizing the student network in knowledge distillation.
Recent works reduce bias in the knowledge provided by the teacher network by weighting~\cite{He2023,Zhang2023} and softening~\cite{He2021} teacher predictions.
However, unbalanced representations and misclassified predictions of the teacher network are still ignored despite their potential to mislead student networks.
Therefore, we propose the KRDistill to rectify the imbalanced feature representations and misclassified predictions from the teacher network, and transfer clear and balanced knowledge to learn a reliable student network.

\section{Implementation Details}

The overall process of our proposed KRDistill is summarized in Algorithm 1.
Before distillation, we calculate the prior mean feature representations generated by the teacher network and obtain the ideal feature representations through Eq. (2).
During the distillation process, the student network learns the balance feature representation rectified by ideal feature representations through Eq. (5) and the precise teacher predictions through Eq. (7).

\begin{algorithm}[h]
    \caption{Knowledge Rectification Distillation}
    \label{alg1}
    \textbf{Input}: A long-tailed training set $\mathcal{D}=\left\{\mathcal{D}_1^{n_1},\mathcal{D}_2^{n_2},...,\mathcal{D}_C^{n_C}\right\} $ containing $C$ categories, a pre-trained teacher network $\mathcal{T}$, a
    randomly initialized student network $\mathcal{S}$, the total epoch $E$.\\
    \textbf{Output}: Parameters of a reliable student network after training.
    \begin{algorithmic}[1] 
        \FOR{$x \in \mathcal{D}$}
        \STATE Calculate the feature representation mean $\left\{\boldsymbol{\mu}_1, \boldsymbol{\mu}_2,..., \boldsymbol{\mu}_C\right\}$ generated by the teacher network.
        \ENDFOR

        \STATE Taking the feature representation mean of the teacher network as the prior, minimize Eq. (2) to obtain the ideal feature representation for each category.

        \FOR{$e$ in 1,2,...,$E$}
            \FOR{$c$ in 1,2,...,$C$}
                \FOR{$\boldsymbol{x} \in \mathcal{D}_c^{n_c}$}
                \STATE Obtain feature representation $\boldsymbol{f}_{\boldsymbol{x}}^\mathcal{T}$ and prediction $\boldsymbol{p}_{\boldsymbol{x}}^\mathcal{T}$ of the teacher network.
                \STATE Modify the $\boldsymbol{f}_{\boldsymbol{x}}^\mathcal{T}$ using Eq. (4).
                \STATE Calculate representation-rectified distillation loss $\mathcal{L}_{RRD}$ using Eq. (5).
                \STATE Modify the $\boldsymbol{p}_{\boldsymbol{x}}^\mathcal{T}$.
                \STATE Calculate logit-rectified distillation loss $\mathcal{L}_{LRD}$ using Eq. (7).
                \STATE Calculate the total loss $\mathcal{L}_{Total}$ of the student network by Eq. (8).
                \ENDFOR
            \ENDFOR
        \STATE Update parameters of the student network by minimizing $\mathcal{L}_{Total}$.
        \ENDFOR
    \end{algorithmic}
\end{algorithm}

\section{Experimental Details}\label{sec41}

\subsection{Datasets} 
Our experiments are conducted on the five public long-tailed datasets to verify the effectiveness of our proposed KRDistill in the long-tailed scenarios.
The number of training examples and imbalance rates of the dataset are summarized in Table~\ref{datasets}.

1) \textbf{CIFAR-LT} is obtained by randomly sampling examples from the original CIFAR dataset~\cite{Krizhevsky2009}, which contains 5,000 images from 10 classes in CIFAR10 and 50000 images from 100 classes in CIFAR100.
We follow the widely used dataset processing method~\cite{Cui2019} to construct the CIFAR10-LT and CIFAR100-LT datasets, setting the imbalance ratios $\rho$ to 100 and 50 in our experiments.

2) \textbf{ImageNet-LT} is a subset of ImageNet~\cite{Deng2009} that follows a Poisson distribution with $\gamma=0.6$, which comprises 1,158K images from 1000 classes. The number of examples in each category in ImageNet-LT exhibits severe imbalance, varying from 1,280 to 5.

3) \textbf{Places365-LT}~\cite{Zhou2017} is the long-tailed variant of Places365. This dataset includes 184K images from 365 categories.
The severe imbalance rate of 4980/5 poses challenges for the visual recognition tasks on the Places365-LT dataset.

4) \textbf{iNaturalist2018}~\cite{VanHorn2018} is a large-scale real-world dataset frequently employed for long-tailed recognition tasks, which contains 437K training images across 8,142 categories with an extreme imbalance rate of 118.8K/16. 
To ensure fair comparisons, we employ the official segmentation method\footnote{https://github.com/visipedia/inat\_comp/blob/master/2018.} of the training and validation sets in our experiments. 

\begin{table}[h]
  \caption{The total number of examples (Num.) and imbalance rates of each long-tail dataset. The imbalance rate $\rho$ represents the ratio between the most frequent and least frequent classes.}
  \begin{tabular}{p{4.3cm}cc}
    \toprule
    Dataset & Num. & Imbalance rate ($\rho$) \\
    \midrule
    CIFAR10-LT~\cite{Krizhevsky2009} & 5K & 50 and 100 \\
    CIFAR100-LT~\cite{Krizhevsky2009} & 50K & 50 and 100\\
    ImageNet-LT~\cite{Deng2009} & 115K & 256 \\
    Places365-LT~\cite{Zhou2017} & 184K & 996 \\
    iNaturalist2018~\cite{VanHorn2018} & 437K & 7425 \\
  \bottomrule
\end{tabular}\label{datasets}
\end{table}

\subsection{Teacher-Student Model Architecture} 
We use various teacher-student architectures with different compression ratios to verify the performance of KRDistill under different compression requirements.
As shown in Table~\ref{Arch}, we conduct experiments using teacher-student architectures with a minimum compression ratio of 56.3\% and a maximum of 99.5\%.

\begin{table}[t]
  \caption{Comparison of parameter quantities (Param.) and compression ratio (Ratio) of different teacher-student architectures in our experiments. The compression ratio is calculated by the ratio of the parameter difference between the teacher and student network to the parameter amount of the teacher network.}
  \begin{tabular}{ccccc}
    \toprule
    \multicolumn{2}{c}{Teacher} & \multicolumn{2}{c}{Student} & \multirow{2}{*}{Ratio}\\
    Model & Param. & Model & Param. & \\
    \midrule
    ResNet-152~\cite{He2016} & 35.5M & ResNet-50~\cite{He2016}  & 15.5M & 56.3\%\\
    ResNet-110~\cite{He2016} & 257.4M  & ResNet-32~\cite{He2016} & 70.4M  & 72.6\%\\
    ResNet-110~\cite{He2016} & 257.4M & ResNet-32~\cite{He2016} & 70.4M  & 72.6\%\\
    ResNet-152~\cite{He2016} & 35.5M & MobileNetV2~\cite{Sandler2018} & 7.9M  & 77.7\%\\
    ResNext-50~\cite{Xie2017} & 15.2M  & ResNet-10~\cite{He2016} & 3.3M  & 78.3\%\\
    ViT-base~\cite{Dosovitskiy2020} & 113.7M  & ViT-tiny~\cite{Dosovitskiy2020} & 5.4M  & 95.3\%\\
    ResNet-110~\cite{He2016} & 257.4M  & ShuffleNetV2~\cite{Ma2018} & 1.4M & 99.5\%\\
  \bottomrule
\end{tabular}\label{Arch}
\end{table}

\subsection{Experimental Settings}

In our experiments, we use cumbersome ResNext-50~\cite{Xie2017}, ResNet-110~\cite{He2016}, and ResNet-152 as pre-trained teacher networks to provide informative knowledge for the training of lightweight student networks (ResNet-50, ResNet-32, ResNet-10, and MobileNetV2~\cite{Sandler2018}).
The comparison of the amount of parameters and calculations between teacher networks and student networks is shown in the supplementary material.
The hyper-parameters, including the weight of the representation-rectified distillation loss, exponential moving average rate, temperature, and the number of hidden layers in MLP are set to 10, 0.8, 2, and 3, respectively. The sensitivity of our proposed method to these hyper-parameters is discussed in Section~\ref{sec44}.

For the CIFAR10-LT and CIFAR100-LT datasets, we train the student network (ResNet-32~\cite{He2016}) for 200 epochs. The batch size is set to 128. 
For experiments on ImageNet-LT, ResNet-10~\cite{He2016} are trained for 180 epochs with a batch size of 256.
For the MobileNetV2 trained on Places365-LT, we set batch size and total epoch to 128 and 90.
For the iNaturalist2018 dataset, we train ResNet-50 for 90 epochs with a batch size of 512.
The Stochastic Gradient Descent~\cite{Bottou2012} optimizer with a momentum of 0.9 is used for experiments on all datasets.
Except for the iNaturalist2018 and Places365-LT dataset, the weight decay is set to 5e-4 and the initial learning rate is set to 0.1.
The weight decay and the initial learning rate are 2e-4 and 0.2 for the iNaturalist2018 dataset and 4e-4 and 0.01 for the Places365-LT dataset.
The cosine scheduler is used to decay the initial learning rate as training progresses.

\section{More verification experiments}

We conduct more verification experiments to verify the generalizability of our proposed KRDistill.
Table~\ref{cifar10} shows the experimental results of using different teacher-student architectures under a high compression ratio on the CIFAR100-LT dataset.
KRDistill significantly outperforms traditional VKD~\cite{hinton2015} and advanced BKD~\cite{Zhang2023} in both convolution-based architecture ShuffleNetV2 and transformer-based architecture ViT-tiny, which proves that our proposed KRDistill is adaptable to different architectures and can train a reliable student network even under high compression rates.

\begin{table}[t]
    \centering
    \caption{The Top-1 classification accuracy (\%) of student networks with different architectures on the CIFAR100-LT with imbalance rates of 100 and 50. Superscripts $\mathcal{T}$ and $\mathcal{S}$ are used to mark the teacher network and student network respectively. The highest accuracy rates are highlighted in bold.}

    \begin{tabular}{p{3cm}cccc}
        \toprule
        \multirow{3}{*}{Method} & \multicolumn{2}{c}{ResNet-110$^{\mathcal{T}}$} & \multicolumn{2}{c}{ViT-base$^{\mathcal{T}}$}\\
        & \multicolumn{2}{c}{ShuffleNet$^{\mathcal{S}}$} & \multicolumn{2}{c}{ViT-tiny$^{\mathcal{S}}$}\\
        & $\rho$ = 50 & $\rho$ = 100 & $\rho$ = 50 & $\rho$ = 100 \\
        \midrule
        CE$^{\mathcal{T}}$ & 51.9 & 46.1 & 38.8 & 34.5 \\
        CE$^{\mathcal{S}}$ & 46.3 & 40.9 & 30.5 & 27.2 \\
        VKD~\cite{hinton2015} & 54.1 & 48.0 & 31.4 & 28.4 \\
        LS~\cite{Sun2024}+DKD~\cite{Zhao2022} & 50.7 & 44.8 & 37.2 & 24.1 \\
        BKD~\cite{Zhang2023}  & 58.6 & 53.6 & 31.4 & 27.7 \\
        KRDistill & \textbf{59.0} & \textbf{54.0} & \textbf{40.3} & \textbf{37.9} \\
        \bottomrule
        
    \end{tabular}
    \label{cifar10}

\end{table}

\section{Parameter-Sensitivity Experiment}\label{sec44}
In this subsection, we analyze the sensitivity of hyper-parameters involved in the KRDistill, including the weight of representation rectified distillation loss $\beta$, the exponential moving average speed $\alpha$,  the temperature $\tau$ in the distillation process, and the number of layers of the multilayer perceptron in Eq. (5).
Here, we employ the ResNet-110 and ResNet-32 as the teacher network and student network, respectively, to train on the CIFAR100-LT dataset with an imbalanced rate of 100.

\subsection{The Weight of Representation-Rectified Distillation}
\begin{figure*}[t]
  \centering
  \includegraphics[trim=0cm 11.5cm 0cm 0cm, clip, width=1.0\textwidth]{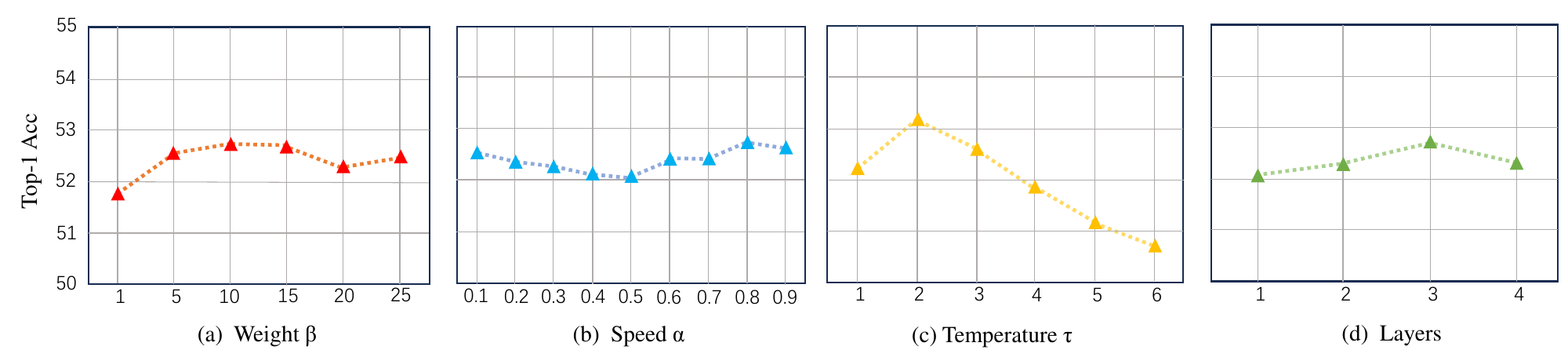}
  \caption{The impact of (a) different representation-rectified distillation loss weight values, (b) exponential moving average speeds, (c) temperature $\tau$, and (d) numbers of hidden layers of MLP on KRDistill performance on CIFAR100-LT with an imbalance rate $\rho$ of 100.}
  \label{fig4}
\end{figure*}

The hyper-parameter $\beta$ is used to weight the feature-based loss components, \textit{i.e.} 
 Representation-Rectified Distillation loss in Eq. (8).
We investigate the impact of different weight values on the performance of KRDistill.
The results in Figure~\ref{fig4} (a) demonstrate the insensitivity of our method to the weight of representation-rectified distillation loss.
Different weight values only result in a maximum 1\% fluctuation in recognition accuracy.
In our method, we set the weight of representation-rectified distillation loss $\beta$ to 10.

\subsection{The Exponential Moving Average Speed.}
To mitigate excessive storage space usage, we employ an exponential moving average to compute the category feature mean in Representation-Rectified Distillation loss:
\begin{equation}\label{EMA}
\left\{
\begin{aligned}
    \boldsymbol{\mu}_{c,k} &= \boldsymbol{f}^{\mathcal{T}}_{c,k}, &&\text{ if } k=1; \\
    \boldsymbol{\mu}_{c,k} &= \alpha\boldsymbol{\mu}_{c,k-1} + \left(1 - \alpha\right)\boldsymbol{f}^{\mathcal{T}}_{c,k}, &&\text{ if } k\in\left\{2, ..., n_c\right\},
\end{aligned}
\right.
\end{equation}
where $k$ represents the example index and $n_c$ is the total number of examples in the $c$-th category,
$\boldsymbol{f}_{c,k}$ is the regularized feature representation of the $k$-th sample in the $c$-th class, 
and $\alpha$ is a hyper-parameter utilized to control the rate of movement. 
Figure~\ref{fig4} (b) shows the impact of different values of hyper-parameter $\alpha$ on the performance of our proposed KRDistill.
It is easy to observe that our proposed method performs stable at different speeds and tends to perform slightly worse when the $\alpha$ is smaller.
In our method, we set $\alpha$ to 0.8 in Equation~\ref{EMA}.

\subsection{The Temperature in Distillation Process.}
The temperature hyperparameter $\tau$ was introduced by Hinton~\textit{et al.}~\cite{hinton2015} to control the smoothness of teacher and student predictions during the distillation process.
An appropriate temperature value can effectively improve the learning effect of the student network.
Figure~\ref{fig4} (c) shows the impact of different temperatures on the performance of the student network trained by our KRDistill.
In our study, we follow BKD~\cite{Zhang2023} to set $\tau$=2 to optimize the distillation effect.

\subsection{the number of layers in the multilayer perceptron.}
In our method, a Multilayer Perceptron (MLP) is used to align the dimensions of student and teacher feature representations.
The number of hidden layers in MLP will affect the alignment effect of feature representation, which in turn affects the learning effect of the student network.
We explore the impact of different numbers of hidden layers on the performance of our proposed KRDistill.
As shown in Figure~\ref{fig4} (d), our method is not sensitive to the number of hidden layers, and changes in the number of hidden layers will only have a weak impact on the performance of KRDistill by up to about 0.6\%.
We set the number of hidden layers in MLP to 3 in our experiments.

\section{Discussion on Computational Complexity}\label{sec45}
Compared with the vanilla knowledge distillation process, our proposed representation-rectified distillation loss requires ideal feature representations as prior knowledge to rectify the imbalanced feature knowledge provided by the teacher network, which entails a slight increase in computational overhead.
Table~\ref{time} shows the extra computation time incurred by our method on different datasets, utilizing an RTX4080 for calculations on small-scale datasets (CIFAR10-LT and CIFAR100-LT) and four RTX4090 on large-scale datasets (ImageNet-LT, Places365-LT, and iNaturalist2018).

For calculating the mean of teacher feature representations, only a single inference pass is conducted on the training dataset using the pre-trained teacher network.
This process typically lasts only tens of seconds, even for large-scale datasets, making it a swift operation in contrast to the training of student networks.

Regarding the generation of the ideal feature representation, the category feature representation mean is utilized as the initialization parameter, and the Stochastic Gradient Descent optimizer is used to optimize with Eq. (2) as the loss function.
The computation time of this process is related to the total number of categories and feature representation dimensions.
Table~\ref{time} shows the calculating time required for optimizing the ideal feature representation of different feature dimensions across multiple datasets.
Even in the iNaturalist2018 dataset containing 437K training images, optimizing 8142$\times$1536 dimensional features for 20,000 epochs only takes around 5 minutes, which only accounts for about 0.56\% of the total training time.

In summary, our method introduces nearly negligible computational overhead when the dataset categories and feature dimensions are limited.
For scenarios with high numbers of categories and feature dimensions, such as 8142$\times$1536 dimension for the iNaturalist2018 dataset, our method only requires a few additional minutes to calculate the feature representation mean and generate the ideal feature dimension. 
This additional time is minimal compared with the overall training duration. Once the model enters the training phase, the computational cost of our method is almost indiscernible from that of vanilla knowledge distillation methods.
\begin{table}[t]
    \centering
    \caption{The computation time introduced by KRDistill and its percentage of training time. Dimen. represents the dimension of the category representation mean. Cal. and Gen. denote the computational time of the calculation of representation means and the generation of the ideal representation, respectively.}
    \begin{tabular}{ccccc}
        \toprule
        Dataset & Dimen. & Cal.  & Gen. \\
        \midrule
        CIFAR100-LT  & 100$\times$64 & 4s(0.22\%) & 3s(0.17\%)  \\
        ImageNet-LT  & 1000$\times$1536 & 52s(0.15\%) & 37s (0.11\%)\\
        Places365-LT  & 365$\times$2048 & 75s(0.35\%) & 34s (0.16\%)\\
        iNaturelist2018  & 8142$\times$1536 & 478s (0.83\%) & 324s (0.56\%) \\
        \bottomrule
    \end{tabular}\label{time}
\end{table}


%% file: _Abstract.tex
Knowledge Distillation (KD) transfers knowledge from a large pre-trained teacher network to a compact and efficient student network, making it suitable for deployment on resource-limited media terminals. However, traditional KD methods require balanced data to ensure robust training, which is often unavailable in practical applications. In such scenarios, a few head categories occupy a substantial proportion of examples. This imbalance biases the trained teacher network towards the head categories, resulting in severe performance degradation on the less represented tail categories for both the teacher and student networks. In this paper, we propose a novel framework called \textbf{K}nowledge \textbf{R}ectification \textbf{Distill}ation (KRDistill) to address the imbalanced knowledge inherited in the teacher network through the incorporation of the balanced category priors. Furthermore, we rectify the biased predictions produced by the teacher network, particularly focusing on the tail categories. Consequently, the teacher network can provide balanced and accurate knowledge to train a reliable student network. Intensive experiments conducted on various long-tailed datasets demonstrate that our KRDistill can effectively train reliable student networks in realistic scenarios of data imbalance.

%% file: _Intro.tex
In recent years, deep learning models with massive parameters have achieved remarkable progress~\cite{duan2022pyskl, lin2023uninext, sun2023mae, li2021replay}.
However, these advanced deep learning models often necessitate massive storage and computational resources, rendering them unsuitable for deployment on small media devices with limited resources.
To address this issue, various model compression techniques have been developed, mainly including network pruning~\cite{lin2017runtime, molchanov2019importance}, parameter quantization~\cite{wu2016quantized,kryzhanovskiy2021qpp}, and knowledge distillation~\cite{hinton2015distilling,tang2023distribution}.
Among these approaches, Knowledge Distillation (KD) is simple and effective, which enhances the performance of compact student networks by mimicking knowledge from a well-trained yet completed teacher network.

\begin{figure}[tb]
  \centering
  \includegraphics[trim=2cm 3cm 1cm 1cm,clip,scale=3,width=0.47\textwidth]{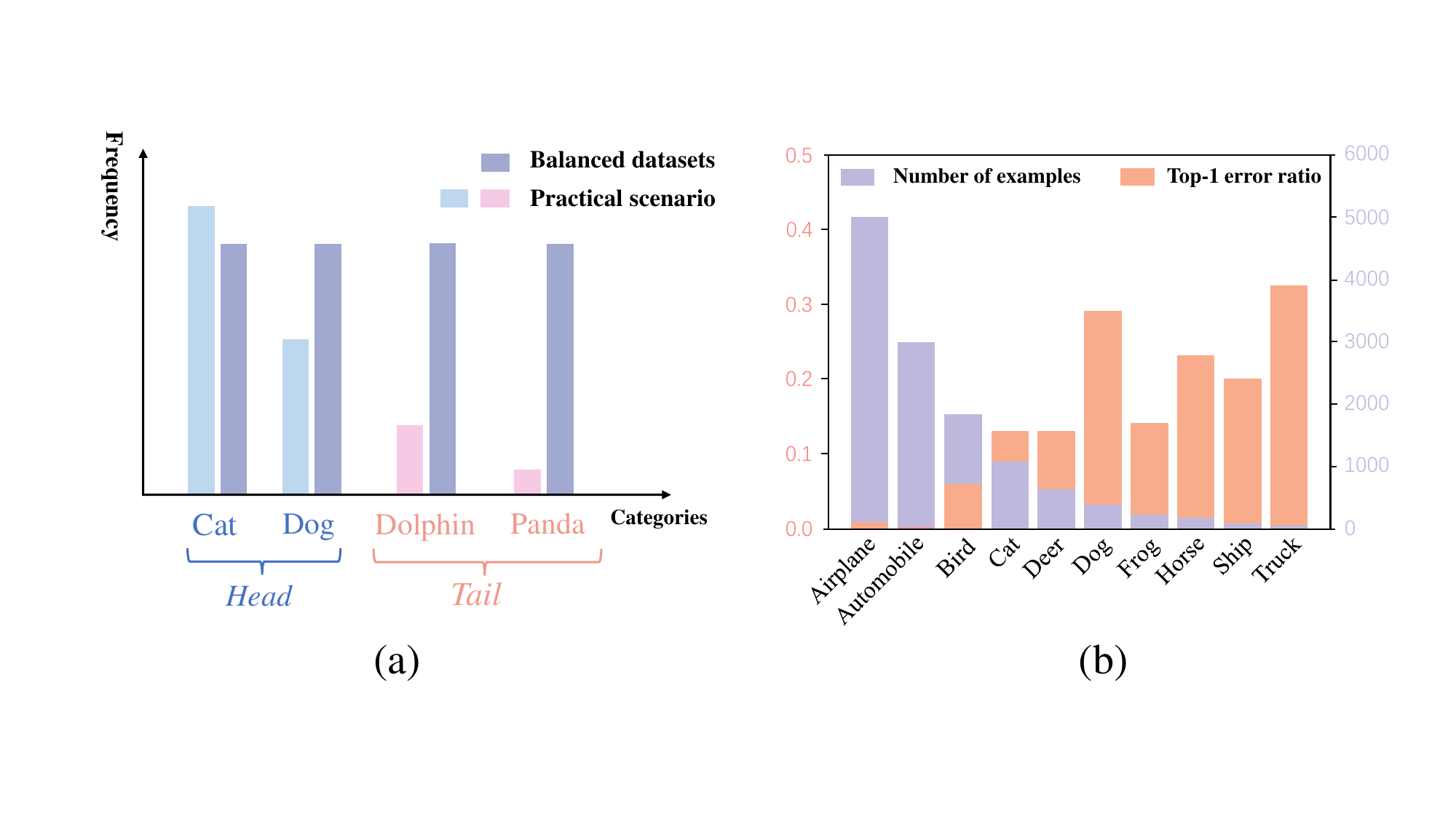}
  \vspace{-3mm}
  \caption{(a) Comparison of example distributions in balanced datasets and long-tailed data in practice scenarios. (b) Top-1 error rate of the teacher network (ResNet-110) per category on the CIFAR10-LT dataset.}
  \label{fig1}
  \vspace{-5mm}
\end{figure}

Conventional KD methods often assume that both the teacher and student networks are trained on meticulously balanced datasets (\textit{e.g.}, CIFAR~\cite{Krizhevsky_2009} and ImageNet~\cite{deng2009imagenet}).
In practice, however, the distribution of real-world data usually tends to be imbalanced, where minority head categories occupy the most examples (such as ``Cat'' and ``Dog'') while the remaining tail categories only have a few examples (``Dolphins'' and ``Panda'') as illustrated in Figure~\ref{fig1}.
In this scenario, the teacher network trained on the imbalanced data will inevitably bias towards the head categories and only achieve poor performance on the tail categories, as shown in Figure~\ref{fig1} (b). 
As a result, the flawed knowledge provided by the teacher network adversely impacts the performance of the student network.

\begin{figure}[tb]
  \centering
  \includegraphics[trim=1.5cm 2cm 2cm 2cm,clip,width=0.47\textwidth]{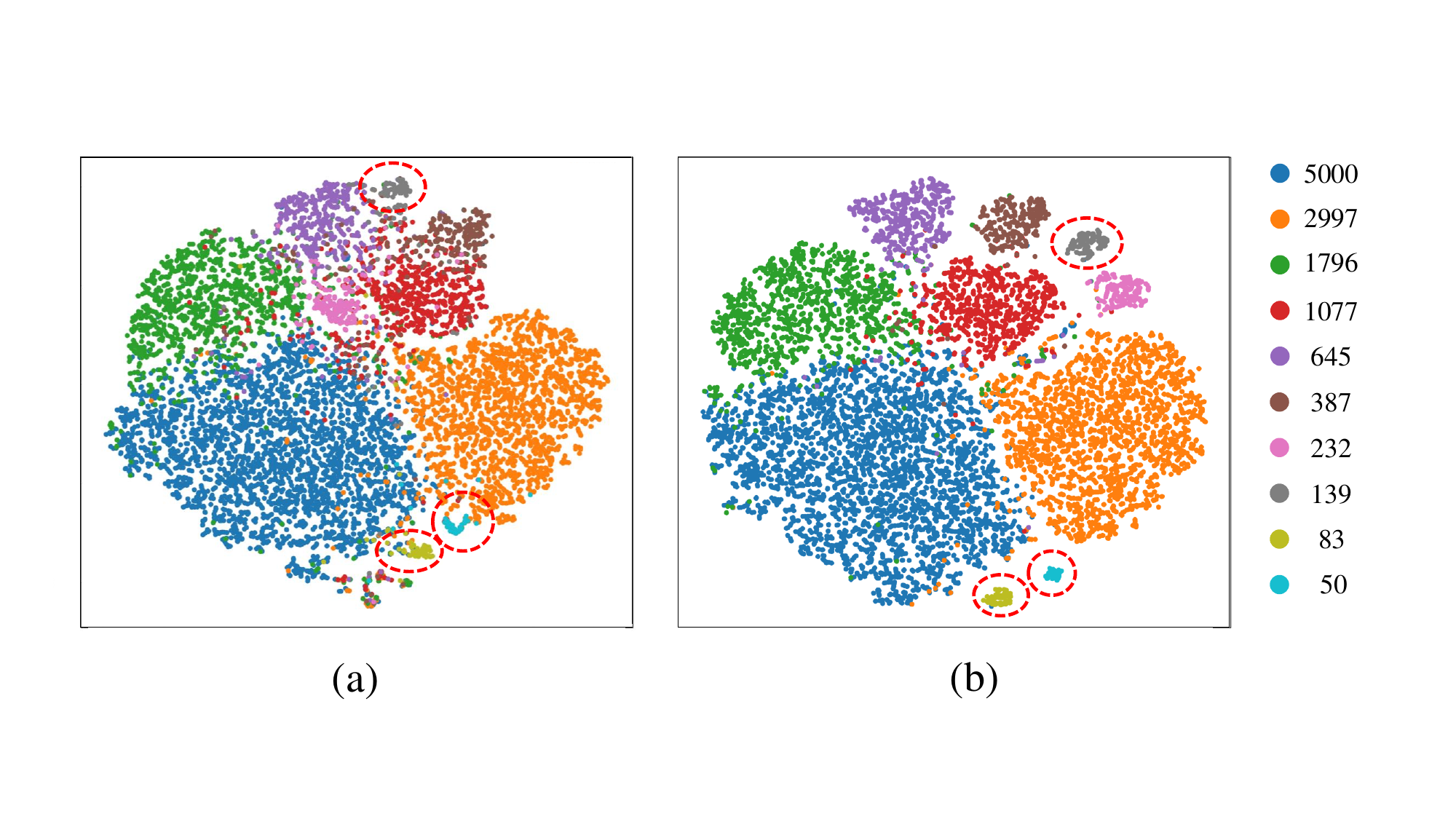}
  \vspace{-3mm}
  \caption{Visualization of (a) feature representations generated by the imbalanced teacher network, (b) modified teacher feature representations using our method on the CIFAR10-LT dataset. The number of examples in each category is marked on the right. 
  }
  \label{tsne}
  \vspace{-5mm}
\end{figure}

Recent advancements in KD have attempted to mitigate the negative impacts of imbalanced long-tailed data.
Zhang~\textit{et al.}~\cite{zhang2023balanced} and He~\textit{et al.}~\cite{he2023joint} reweight the logits of the teacher network to balance the gradient contributions between the head classes and tail classes.
He~\textit{et al.}~\cite{he2021distilling} propose the temperature rise mechanism to smooth the predictions of the teacher network.
Iscen~\textit{et al.}~\cite{iscen2021class} ensemble the knowledge of multiple teacher networks to provide robust knowledge to the student network.
However, these methods often neglect two critical issues:
1) \textit{Imbalanced representations}: the representations of the imbalanced teacher network are biased toward the head categories and exhibit unclear class boundaries between the head and tail categories (Figure~\ref{tsne} (a)), which fail to provide reliable guidance for a student network;
2) \textit{Error accumulation}: in the long-tailed scenarios, the teacher network tends to misclassify examples from tail categories (Figure~\ref{fig1} (b)), which in turn misleads the student network and further hurts the performance of tail categories.

To solve the above problems, we propose a novel knowledge distillation framework to train compact student networks on the imbalanced dataset, termed \textbf{K}nowledge \textbf{R}ectification \textbf{Distill}ation (KRDistill).
Specifically, we propose a representation-rectified distillation loss to clarify the boundary between categories within the dataset. Therefore, the teacher network can provide balanced feature representations to the student network. Meanwhile, for the misclassified knowledge of tail categories, we propose a logit-rectified distillation to adaptively correct the misclassifications caused by the teacher network and transfer the rectified category predictions to the student network. Thanks to the balanced representations and precise predictions from the teacher network, our proposed KRDistill can successfully train reliable and compact student networks on long-tailed datasets with serious class imbalances. In summary, the contributions of this work are as follows:
\begin{itemize}
\item We explore a novel model compression scenario for learning student networks on imbalanced long-tailed data and design the KRDistill to transfer the balanced and precise knowledge from the teacher to the student network.
\item We propose a representation-rectified distillation loss and logit-rectified distillation loss to rectify the imbalanced representations and imperfect predictions of the teacher networks, respectively, and then transfer this valuable rectified knowledge to improve the performance of the student network.
\item Intensive experiments demonstrate that our method can outperform existing state-of-the-art KD works tailored for long-tailed scenarios.

\end{itemize}

%% file: _Method.tex
\begin{figure*}
  \centering
  \hspace{4cm}
  \includegraphics[trim=25 25 0 20,clip, width=1\textwidth]{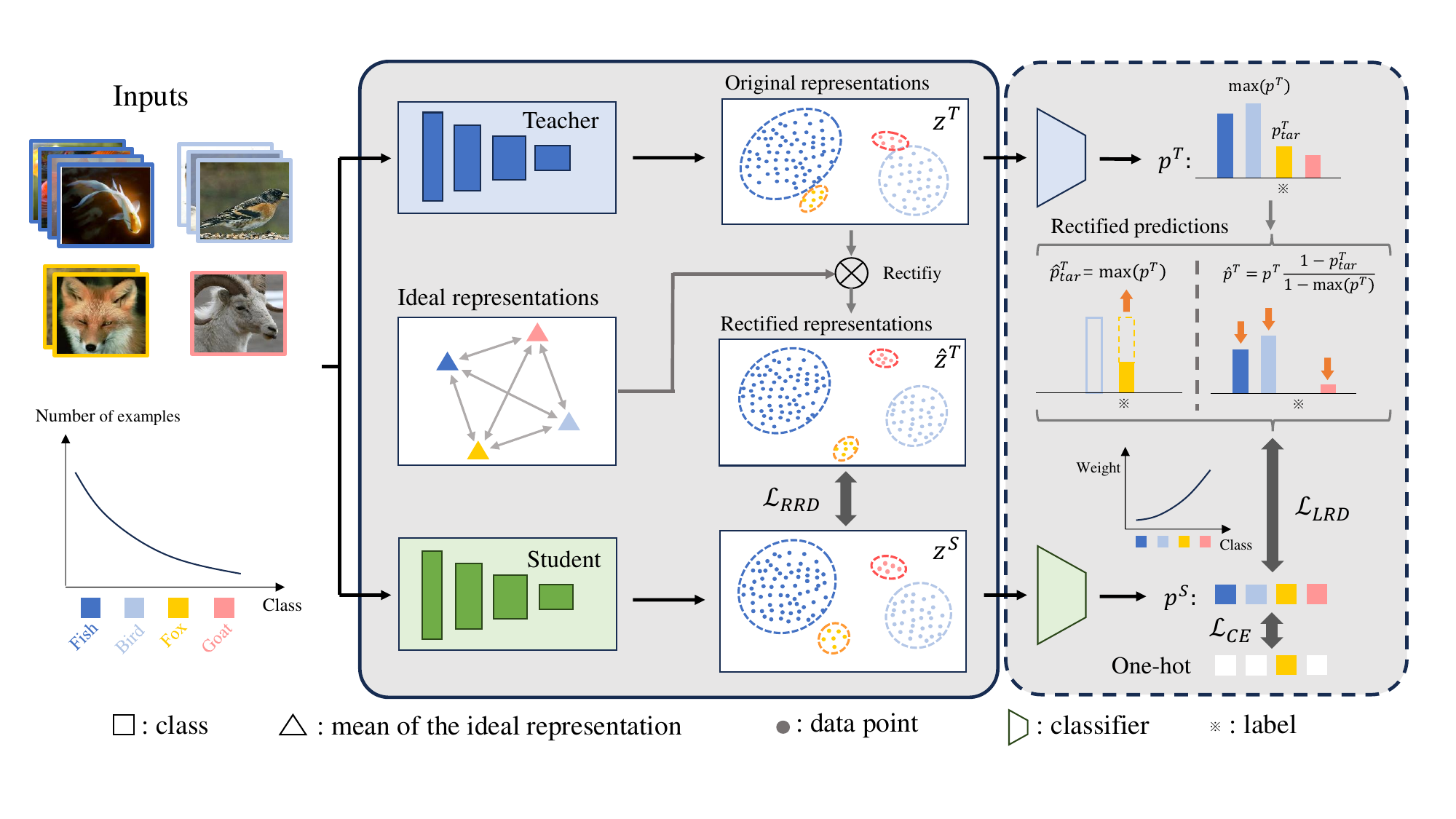}
  \vspace{-8mm}
  \caption{The framework diagram of the proposed Knowledge Rectification Distillation. 
  Ideal feature representations rectify imbalanced teacher features, transferring knowledge of representation with clear class boundaries to the student network.
 Misclassified teacher predictions are adaptively corrected and rebalanced, preventing potential misleading of the student network by imbalanced teacher prediction knowledge.
  }
  \label{_frame}
  \vspace{-5mm}
\end{figure*}


\subsection{Preliminary}\label{sec31}

Knowledge distillation encourages a lightweight student network $\mathcal{S}$ to mimic a well-trained large teacher network $\mathcal{T}$.
Given a training set $\mathcal{D}=\left\{\mathcal{D}_1,\mathcal{D}_2,...,\mathcal{D}_C\right\} $ containing $C$ categories, where $\mathcal{D}_c=\{(\boldsymbol{x}_{i}, c)\}_{i=1}^{n_c}$ represents the $c$-th category containing $n_c$ examples.
KD methods~\cite{chen2021distilling,zhao2022decoupled} transfer the feature representations and predictions of the teacher network to train the student network.
Specifically, the transferring is achieved by minimizing the feature distance and Kullback-Leibler divergence of the predictions between the student and teacher networks:
\begin{equation}
\mathcal{L}_{KD} = \frac{1}{N} \sum_{i=1}^{N}\left(Dis\left(\boldsymbol{f}_{i}^\mathcal{S}, \boldsymbol{f}_{i}^\mathcal{T}\right) + \boldsymbol{p}_{i}^\mathcal{S}log\left(\frac{\boldsymbol{p}_{i}^\mathcal{S}}{\boldsymbol{p}_{i}^\mathcal{T}}\right)\right),
\end{equation}
where $\boldsymbol{f}_{i}$ and $\boldsymbol{p}_{i}$ represent the feature and prediction corresponding to the $i$-th example, respectively.
$Dis\left(\cdot,\cdot\right)$ represents a metric function to estimate the discrepancy between features.
$N=\sum_{c=1}^{C}n_c$ defines the total number of examples.

Traditional distillation methods assume the example distributions across categories are approximately equal.
However, in real scenarios, data distribution often exhibits long-tail characteristics as illustrated in Figure~\ref{fig1} (a), 
which leads to biased representations and predictions from the teacher network, particularly affecting the performance of the student network on tail classes. 
To tackle this issue, this paper proposes the representation-rectified distillation and logit-rectified distillation methods to correct biased knowledge as shown in Figure~\ref{_frame}.

\subsection{Representation-Rectified Distillation}\label{sec32}

Ideally, feature representations for $C$ classes should converge to a $C$-dimensional regular simplex in geometric space~\cite{li2022targeted,zhu2022balanced}, ensuring distinct class boundaries. 
However, in long-tailed scenarios, dominant head categories blur these boundaries in the teacher network, as shown in Figure~\ref{tsne} (a), leading to a suboptimal performance of the student network. 
To mitigate this, we propose a representation-rectified distillation to refine the teacher's feature representations.

Formally, we denote the mean values of the feature representations generated by the teacher network for $C$ categories as $\{\boldsymbol{\mu}_c\}_{c=1}^{C}$.
Since the teacher network is pre-trained, the representation means of $C$ categories can be obtained before distillation begins.
Taking the category representation means as priors, we follow ~\cite{li2022targeted} to obtain the ideal feature representation means $\{\boldsymbol{\hat{\mu}}_c\}_{c=1}^{C}$ by minimizing the following function:
\begin{equation}\label{eq4}
\hat{\mu}:=\argmin_{\mu}\frac{1}{C}\sum_{i=1}^C\log\sum_{j=1}^Ce^{\boldsymbol{\mu}_i^{\top}\cdot \boldsymbol{\mu}_j},
\end{equation}
where ``$\top$" denotes the transpose operation.
Then, we rectify the imbalanced feature representation from the teacher by moving the features toward the ideal feature representation of the corresponding class:
\begin{equation}
\hat{\boldsymbol{F}}_c^\mathcal{T} = \left\{\boldsymbol{f}^{\mathcal{T}}_{c, k} + \boldsymbol{\hat{\mu}}_c\right\}_{k=1}^{n_c} .
\end{equation}

Since the tail classes have few examples in the long-tailed dataset, it is more difficult for the model to learn a well-distinguishable category representation for the tail classes compared to the head classes.
Therefore, we combine the re-weight method to control the degree of rectification:
\begin{equation}\label{eq_our_feat}
\hat{\boldsymbol{F}}_c^\mathcal{T} = \left\{\boldsymbol{f}^{\mathcal{T}}_{c, k} + w_c\boldsymbol{\hat{\mu}}_c\right\}_{k=1}^{n_c} , \text{where }w_c = \frac{C}{n_c\sum_{i=1}^{C}\left(\frac{1}{n_i}\right)}.
\end{equation}
Finally, based on the rectified representations of the teacher network, the student network learns balanced feature representation knowledge by minimizing the \textbf{r}epresentation-\textbf{r}ectified \textbf{d}istillation loss $\mathcal{L}_{RRD}$:
\begin{equation}\label{eq_rrd}
\mathcal{L}_{RRD} = \frac{1}{N} \sum_{c=1}^{C}\left\|MLP\left(\boldsymbol{F}_{c}^\mathcal{S}\right), \hat{\boldsymbol{F}}_{c}^\mathcal{T}\right\|_2.
\end{equation}
where MLP$\left(\cdot\right)$ represents a multilayer perceptron used to align the dimensions of student features to teacher features.
$\left\|\cdot, \cdot\right\|_2$ is the Euclidean distance, used to measure the distance between two feature representations.

\subsection{Logit-Rectified Distillation}\label{sec33}
In the long-tailed scenarios, the trained model will unavoidably overfit the head categories while underfitting the tail categories. Therefore, the teacher network is prone to produce misclassified predictions, especially for tail categories, as depicted in Figure~\ref{fig1} (b). Furthermore, transmitting such misclassifications to the student network will lead to error accumulation, resulting in serious performance degradation. To mitigate the error accumulation during the knowledge distillation process and ensure a reliable student network, we propose a logit-rectified distillation to correct and balance teacher predictions.

For the prediction $\boldsymbol{p}^\mathcal{T}$ made by the teacher network, we divide $\boldsymbol{p}^\mathcal{T}$ into target prediction $p^{\mathcal{T}}_{tar}$ and non-target prediction $\boldsymbol{p}^{\mathcal{T}}_{ntg}$ according to the corresponding ground-truth label~\cite{zhao2022decoupled}.
Apparently, the maximum value in the misclassified prediction probability vector is not equal to the target class prediction, that is, $max(\boldsymbol{p}^\mathcal{T})\neq p_{tar}$.
We first determine the correctness of the teacher prediction by simply assigning the maximum prediction value among the wrong predictions to the target category: $\hat{p}_{tar}^\mathcal{T} = max\left(\boldsymbol{p}^\mathcal{T}\right)$.
Then, we introduce an adaptive penalty factor $\gamma$ to uniformly penalize non-target prediction to maintain the correlation between non-target classes from being uncontrollably destroyed: $\boldsymbol{\hat{p}}^\mathcal{T}_{ntg} = \frac{\boldsymbol{p}_{ntg}^\mathcal{T}}{\gamma}$.

Considering the stability of the training, the value of $\gamma$ should make the sum of the rectified teacher prediction of 1.
Therefore, $\gamma$ is adaptively determined based on the $\boldsymbol{p}^\mathcal{T}$: 
\begin{equation}
\gamma = \frac{1-max\left(\boldsymbol{p}^\mathcal{T}\right)}{1-p_{tar}^\mathcal{T}}.
\end{equation}

Note that in the case of a correct teacher prediction, where $max(\boldsymbol{p}^\mathcal{T}) = p_{tar}^\mathcal{T}$, the value of $\gamma$ is 1, implying that no transformation is applied to the prediction. 
Finally, we follow~\cite{zhang2023balanced} weighted revised teacher prediction $\hat{p}^\mathcal{T}$ to obtain a \textbf{l}ogit-\textbf{r}ectified \textbf{d}istillation loss:
\begin{equation}\label{lld}
\mathcal{L}_{LRD} = \frac{1}{N} \sum_{c=1}^{C}\sum_{i=1}^{n_c}\left(w_c\boldsymbol{\hat{p}}_{\boldsymbol{x}_i}^\mathcal{S}log\left(\frac{w_c\boldsymbol{\hat{p}}_{\boldsymbol{x}_i}^\mathcal{S}}{\boldsymbol{p}_{\boldsymbol{x}_i}^\mathcal{T}}\right)\right).
\end{equation}

\subsection{Implementation Details}
\label{sec34}
The total objective loss function of the student network consists of three components: 
\begin{equation}\label{eq_total}
\mathcal{L}_{Total} = \mathcal{L}_{CE}+\mathcal{L}_{LRD}+\beta\mathcal{L}_{RRD},
\end{equation}
where $\mathcal{L}_{CE}$ is the cross-entropy loss to measure the distance between the predictions of the student network and the ground-truth labels, $\beta\textgreater0$ is the hyper-parameter that balances the loss component $\mathcal{L}_{RRD}$ of representation-rectified distillation, and the parameter sensitivity of $\beta$ is analyzed in supplementary materials.
The overall process of our proposed KRDistill is summarized in supplementary materials.



%% file: _Experiments.tex

\subsection{Datasets and Experimental Settings}\label{sec41}

Our experiments are conducted on the five public long-tailed datasets, including CIFAR10-LT~\cite{Krizhevsky_2009}, CIFAR100-LT~\cite{Krizhevsky_2009}, ImageNet-LT~\cite{deng2009imagenet}, Place365-LT, and iNaturalist2018~\cite{van2018inaturalist}.
Details of these datasets and experimental settings are provided in the supplementary material\footnote{https://arxiv.org/submit/5844904/view}.





\begin{table*}[thb]
\setlength{\tabcolsep}{2.5mm}
    \centering
    \caption{The Top-1 accuracy in three large-scale long-tail datasets. LT. indicates long-tail visual recognition methods. Lgt. KD and Feat. KD represents logit-based and feature-based distillation methods, respectively. The best results are highlighted in bold.}
    \begin{tabular}{ccccccccccccccc}
        \toprule
        \multirow{2}{*}{Type} & \multirow{2}{*}{Method} & \multicolumn{4}{c}{ImageNet-LT ($\rho=256$)}& \multicolumn{4}{c}{Places365-LT ($\rho=996$)} & \multicolumn{4}{c}{iNaturalist2018 ($\rho=7425$)}\\
         & & Head & Medium & Tail & All  & Head & Medium & Tail & All& Head & Medium & Tail & All\\
         
        \midrule
        
        \multirow{2}{*}{Base}& Teacher & 67.9 & 41.9 & 13.2 & 48.0 & 45.6 & 26.9 & 9.1 & 30.2 & 76.4 & 67.9 & 59.9 & 65.6\\
        & Student & 60.1 & 27.8 & 4.4 & 37.0 & 29.3 & 13.8 & 0.4 & 20.3 & 72.2 & 63.1 & 57.4 & 61.8\\
        \midrule
        \multirow{4}{*}{LT.}& CB~\cite{cui2019class} & - & - & - & 37.4 & - & - & - & 25.3 & 47.1 & 54.1 & 53.3 & 53.1 \\
        & AREA~\cite{chen2023area} & 55.7 & 24.8 & 3.5 & 33.8 & 38.0 & 13.2 & 0.5 & 19.7 & - & - & - & 68.4 \\
        & BALMS~\cite{ren2020balanced} & 50.3 & 39.5 & 25.3 & 41.8 & 29.0 & 20.5 & 3.4 & 20.2 & 57.4 & 59.5 & 61.2 & 60.0\\
        & BBN~\cite{zhou2020bbn} & - & - & - & 41.2 & - & - & - & - & 49.4 & 70.8 & 65.3 & 66.3 \\
        \midrule
        \multirow{4}{*}{Lgt. KD}& VKD~\cite{hinton2015distilling} & 61.0 & 26.5 & 3.0 & 36.3 & 44.2 & 18.9 & 2.2 & 24.7 & 75.8 & 66.1 & 58.5 & 64.1\\
        & LS~\cite{sun2024logit}+DKD~\cite{zhao2022decoupled} & 61.3 & 25.1 & 4.1 & 36.2 & 42.4 & 0.2 & 0.03 & 23.4 & 76.3 & 64.8 & 56.1 & 62.5\\
        & CTKD~\cite{li2023curriculum} & 57.8 & 25.9 & 2.8 & 35.1 & 33.4 & 7.3 & 0.04 & 15.3 & 73.6 & 60.9 & 49.9 & 57.8\\
        & BKD~\cite{zhang2023balanced} & 57.8 & 36.4 & 20.8 & 42.5 & 40.8 & 27.6 & 14.7 & 29.8 & 71.6 & 68.0 & 68.0 & 68.4\\
        \midrule
        \multirow{3}{*}{Feat. KD}& ReviewKD~\cite{chen2021distilling} & 59.0 & 27.4 & 3.6 & 36.3 & 35.9 & 8.8 & 0.1 & 16.8 & 76.5 & 65.9 & 57.4 & 63.6 \\
        & SimKD~\cite{chen2022knowledge} & - & - & - & 33.4 & 31.3 & 6.1 & 0.6 & 14.1 & - & - & - & 62.8 \\
        & CAT\_KD~\cite{guo2023class} & 54.8 & 20.9 & 1.7 & 31.3 & 42.0 & 12.3 & 0.3 & 20.6 & - & - & - & 65.0 \\
        \midrule
        Our& KRDistill & 57.9 & 36.9 & 21.7 & \textbf{42.9} & 41.3 & 27.6 & 15.0 & \textbf{30.1} & 72.2 & 68.6 & 68.4 & \textbf{68.9}\\
        \bottomrule
    \end{tabular}\label{places}
    \vspace{-5mm}
    
\end{table*}

\begin{table}[t]
\setlength{\tabcolsep}{2.7mm}
    \centering
    \caption{The Top-1 accuracy (\%) of ResNet-32 on the CIFAR10-LT and CIFAR100-LT datasets with imbalance rates of 100 and 50. The highest accuracy rates are highlighted in bold.}

    \begin{tabular}{cccccc}
        \toprule
        \multirow{2}{*}{Type} & \multirow{2}{*}{Method} & \multicolumn{2}{c}{CIFAR10-LT} & \multicolumn{2}{c}{CIFAR100-LT} \\
        & & $\rho$=100 & $\rho$=50 & $\rho$=100 & $\rho$=50 \\
        \midrule
        \multirow{2}{*}{Base} & Teacher   & 78.2 & 83.9 & 46.1 & 51.9     \\
        & Student   & 74.8 & 79.7 & 40.9 & 46.3     \\
        \midrule
        \multirow{4}{*}{LT}& CB~\cite{cui2019class} & 74.6 & 79.3 & 39.6 & 45.3\\
        & BBN~\cite{zhou2020bbn}  & 79.8 & 82.2 & 42.6 & 47.0\\
        & BALMS~\cite{ren2020balanced} & 84.9 & - & 50.8 & -\\
        & AREA~\cite{chen2023area} & 78.9 & 82.7 & 48.8 & 51.8\\
        \midrule
        \multirow{5}{*}{KD} & VKD~\cite{hinton2015distilling}   & 80.3 &   84.3 & 46.0 & 51.2\\
        &  LS~\cite{sun2024logit}+DKD~\cite{zhao2022decoupled}   & 78.6 &   83.9 & 45.6 & 50.6\\
        & JWAFD~\cite{he2023joint}  & 85.2 & 87.8 & 51.1 & 55.8      \\
        & BKD~\cite{zhang2023balanced}  & 85.3 & 87.8 & 51.7 & 56.0\\
        \midrule
        Our & KRDistill & \textbf{86.2} &   \textbf{88.2} & \textbf{52.7} & \textbf{56.8}  \\
        \bottomrule
    \end{tabular}
    \label{cifar10}
    \vspace{-5mm}

\end{table}

\subsection{Comparison Experiments}
We compare our proposed KRDistill with existing representative works in long-tailed identification, including CB~\cite{cui2019class}, BBN~\cite{zhou2020bbn}, BALMS~\cite{ren2020balanced}, AREA~\cite{chen2023area}; as well as logit-based knowledge distillation (Lgt. KD) methods, including Vanilla Knowledge Distillation (VKD)~\cite{hinton2015distilling}, LS~\cite{sun2024logit}+DKD~\cite{zhao2022decoupled}, CTKD~\cite{li2023curriculum},  BKD~\cite{zhang2023balanced}, and feature-based knowledge distillation (feat. KD) methods, including ReviewKD~\cite{chen2021distilling}, SimKD~\cite{chen2022knowledge}, CAT-KD~\cite{guo2023class}, JWAFD~\cite{he2023joint}, where BKD and JWAFD are
advanced knowledge distillation works in long-tailed scenarios.
As shown in Tables~\ref{places} and Tables~\ref{cifar10}, our method shows consistently state-of-the-art performance in five datasets with different imbalance rates $\rho$, which proves that our proposed KRDistill can effectively suppress the impact of data imbalance and train a reliable student network even in scenarios with severe data imbalance.

\begin{table}[t]
    \centering
    \setlength{\tabcolsep}{2.7mm}
    \caption{Ablation experiment results of our proposed RRD and LRD on the CIFAR100-LT dataset with an imbalanced rate of 100. Acc represents the Top-1 recognition accuracy rate.}
    \begin{tabular}{p{2.8cm}p{1.5cm}p{1.5cm}p{1cm}}
        \toprule
        Method & RRD & LRD & Acc \\
        \midrule
        CE   &  &   & 40.9        \\
        VKD~\cite{hinton2015distilling}   &  &  & 46.0        \\
        KRDistill & $\checkmark$ &   &46.4    \\
        KRDistill &   & $\checkmark$ & 52.0   \\
        KRDistill & $\checkmark$ & $\checkmark$  &52.7    \\
        \bottomrule
    \end{tabular}\label{abl}
    \vspace{-5mm}
\end{table}

\subsection{Ablation Study}\label{sec43}
we examine the contribution of Representation-Rectified Distillation (RRD) loss and Logit-Rectified Distillation (LRD) loss on CIFAR100-LT with an imbalanced rate of 100.
As shown in Table~\ref{abl}, in comparison with cross-entropy (CE) loss, the incorporation of the RRD loss achieves a 5.5\% improvement in the performance of the student network, which can be attributed to the guided of balance feature representations provided by RRD.
Compared with VKD, only using LRD loss can also bring a significant performance improvement of 6.0\% to the student network.
This improvement thanks to LRD loss rectifies misclassified and imbalanced category knowledge from the teacher classifier.

\vspace{-1.5mm}

%% file: _Conclusion.tex
In this paper, we tackle the novel and challenging scenario of learning the student network on the practice of long-tailed data with serious class imbalance. Specifically, to train reliable student networks, our proposed novel KRDistill mainly employs two key operations. First, representation rectification adjusts the imbalanced feature representations of the teacher network towards ideal feature representations. This adjustment enhances the knowledge transfer process, particularly in cases where class boundaries are distinct, enabling effective learning by the student network. Second, logit rectification corrects and rebalances misclassified teacher predictions resulting from data imbalance. This correction process ensures that unbiased category knowledge is provided to the student model. Our experimental evaluations on five long-tailed datasets demonstrate that our proposed KRDistill can train a satisfactory student network in the long-tailed scenarios, thus exhibiting state-of-the-art performance.